\documentclass[journal]{IEEEtran}

\usepackage{cite}
\usepackage[pdftex]{graphicx}
\usepackage{multirow}
\usepackage{booktabs}
\usepackage{comment}

\ifCLASSINFOpdf
\else
\fi
\usepackage{amsmath}
\usepackage{algorithm, algpseudocode}
\ifCLASSOPTIONcompsoc
  \usepackage[caption=false,font=normalsize,labelfont=sf,textfont=sf]{subfig}
\else
  \usepackage[caption=false,font=footnotesize]{subfig}
\fi
\usepackage{amsmath}
\DeclareMathOperator{\arctantwo}{arctan2}

\begin{document}
%
\title{Deep Reinforcement Learning for Concentric Tube Robot Path Following}
%
%
%
\author{Keshav~Iyengar, ~\IEEEmembership{Graduate Student,~IEEE} %
        Sarah~Spurgeon,~\IEEEmembership{Fellow,~IEEE} %
        Danail~Stoyanov~\IEEEmembership{Senior Member,~IEEE}%
\thanks{This work was supported by the Wellcome/EPSRC Centre for Interventional and Surgical Sciences (WEISS) at UCL (203145Z/16/Z), EPSRC (EP/P027938/1, EP/R004080/1). Danail Stoyanov is supported by a Royal Academy of Engineering Chair in Emerging Technologies (CiET1819\textbackslash2\textbackslash 36) and an EPSRC Early Career Research Fellowship (EP/P012841/1). For the purpose of open access, the author has applied a CC BY public copyright license to any accepted manuscript version arising from this submission.}
\thanks{Keshav Iyengar and Danail Stoyanov are with the Wellcome/ EPSRC Centre for Interventional and Surgical Sciences (WEISS), University College London, London W1W 7EJ, UK. Sarah Spurgeon is with the Department of Electronic and Electrical Engineering, University College London, London W1W 7EJ, UK.
        {\tt\small keshav.iyengar@ucl.ac.uk}}%
        }

%
%

\markboth{This work has been accepted for publication in Transactions on Medical Robotics and Bionics.}%
{}
%



\maketitle

\begin{abstract}
    As surgical interventions trend towards minimally invasive approaches, Concentric Tube Robots (CTRs) have been explored for various interventions such as brain, eye, fetoscopic, lung, cardiac, and prostate surgeries. Arranged concentrically, each tube is rotated and translated independently to move the robot end-effector position, making kinematics and control challenging. Classical model-based approaches have been previously investigated with developments in deep learning-based approaches outperforming more classical approaches in both forward kinematics and shape estimation. We propose a deep reinforcement learning approach to control where we generalize across  two to four systems, an element not yet achieved in any other deep learning approach for CTRs. In this way, we explore the likely robustness of the control approach. Also investigated is the impact of rotational constraints applied on tube actuation and the effects on error metrics. We evaluate inverse kinematics errors and tracking errors for path-following tasks and compare the results to those achieved using  state-of-the-art methods. Additionally, as current results are performed in simulation, we also investigate a domain transfer approach known as domain randomization and evaluate error metrics as an initial step toward hardware implementation. Finally, we compare our method to a Jacobian approach found in the literature.
\end{abstract}

\begin{IEEEkeywords}
Kinematics, Reinforcement Learning, Concentric Tube Robots
\end{IEEEkeywords}

%
\IEEEpeerreviewmaketitle

\section{Introduction}

\begin{figure}[tb]
      \centering
     \subfloat[]{\includegraphics[width=\linewidth]{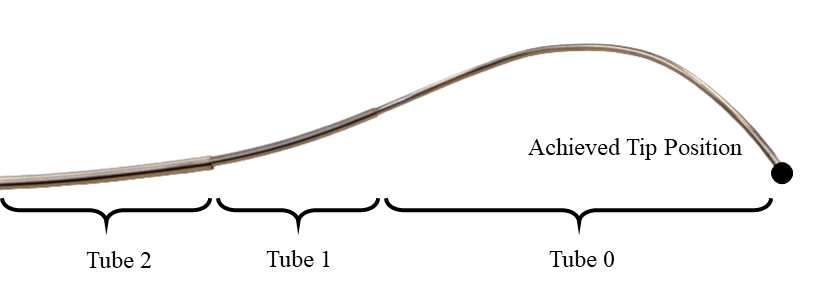}\label{fig:real_ctr}}\\
      \subfloat[]{\includegraphics[width=\linewidth]{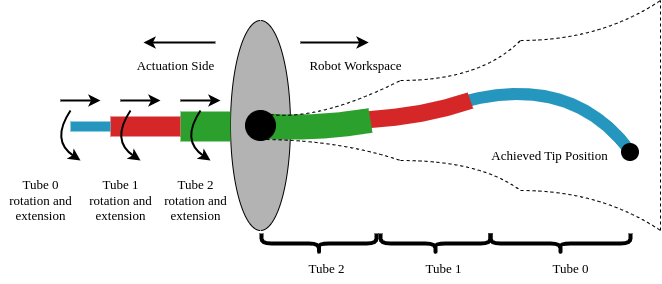}\label{fig:illustrated_ctr}}
      \caption{A real set of concentric tubes \cite{gilbert2015} \protect\subref{fig:real_ctr} and an equivalent illustration with tube actuation, workspace, and tip position.}
      \label{fig:ctr_real_and_illustrated}
\end{figure}

\IEEEPARstart{C}{oncentric} tube robots (CTRs) are a class of continuum robots that depend on the interactions between neighboring, concentrically aligned tubes to produce the curvilinear shapes of the robot backbone \cite{dupont2012}. The main application of these unique robots is that of minimally invasive surgery (MIS), where most of the developments for CTRs have been focused. MIS has trended towards semi-autonomous and autonomous robotic surgery to improve surgical outcomes \cite{demomi2022, dettorre2021}. Due to the confined workspaces and resulting extended learning times for surgeons in MIS, dexterous, compliant continuum robots such as CTRs have been under development in preference to the mechanically rigid and limited degrees-of-freedom (DOF) robots used in interventional medicine today. The pre-curved tubes of CTRs, which are sometimes referred to as active cannulas or catheters, are manufactured from super-elastic materials such as Nickel-Titanium alloys with each tube nested concentrically \cite{dupont2009}. From the base, the individual tubes can be actuated through extension, as seen in Fig. \ref{fig:ctr_real_and_illustrated}, which results in the bending and twisting of the backbone as well as access to the surgical site through the channel and robot end-effector.

CTRs are motivated clinically for use in the brain, lung, cardiac, gastric, and other surgical procedures where tool miniaturization and non-linear tool path trajectories are beneficial \cite{alfalahi2020}. Particularly, they have been investigated as steerable needles and surgical manipulators. As steerable needles, they substitute traditional steerable needles with higher precurvature and dexterity. They are also actuated with a follow-the-leader approach, where every point along the robot's backbone traces the same path as the tip. As surgical manipulators, the benefit of using CTRs is the large number of design parameters that allow for patient-specific and procedure-specific CTR designs through optimization for surgical requirements and design heuristics \cite{bedell2011, burgner2013}. In general, however, CTRs, with their increased DOFs and miniaturization potential, may be beneficial with their flexibility to reach a larger section of a surgical site with a working channel through the tubes for irrigation, ablation, or other tools. Thus, it is key to control the robot tip position to desired Cartesian points in the robot workspace accurately. However, due to tube interactions, modeling, and control are challenging. Position control for CTRs has relied on model development, and although a balance between computation and accuracy has been reached in the literature \cite{gilbert2016}, there remain issues such as performance in the presence of tube parameter discrepancies and the impact of unmodelled physical phenomena such as friction and permanent plastic deformation. This motivates the development of an end-to-end model-free control framework for CTRs.

One such model-free framework for robotic control that is gaining popularity is reinforcement learning (RL), a paradigm of machine learning that deploys an agent to output an action that interacts with an environment \cite{sutton2018}. The environment then processes this action and returns a new state and, depending on the task, a reward signal. One parallel of RL is that of a control system. The agent is equivalent to the controller, the actions are equivalent to the control actions, the state is equivalent to any measurable signal from the plant, the reward is equivalent to a performance metric (eg. minimize steady-state error), and lastly, the learning algorithm is equivalent to the adaptive mechanism of the controller. Deep reinforcement learning (DeepRL) combines deep learning and RL and allows for high-dimensional states and actions traditionally not available to RL algorithms or classical control. In this work, we expand on the previously published literature on utilizing DeepRL to control CTRs \cite{iyengar2021}. Specifically, the aim is to control the end-effector Cartesian robot tip position with a DeepRL agent by means of actions that represent changes in joint values. The state includes the desired goal at the current step, allowing for more complex control tasks such as path following as well as inverse kinematics.

In \S \ref{sec:related_work}, we review the Jacobian approach \cite{rucker2010} as well as the state-of-the-art inverse kinematics error metrics. The model reviewed is also used in simulation for training and evaluation for our DeepRL method. In \S \ref{sec:methods}, the main components of our DeepRL approach for CTRs are described. Improvements such as exploring constrained and constraint-free tube rotation as well as generalizing the policy to accommodate multiple CTR systems are introduced. For tube rotation, we investigate how constraining the rotational degree of freedom for CTRs affects overall error metrics. In other deep learning methods, tube rotations are often constrained during data collection. For example, in \cite{grassmann2018}, a restricted rotation joint space of $-\pi/3$ to $\pi/3$. For a neural network approach, this may not cause issues however, in a timestep-based exploration method like DeepRL, such constraints may hinder exploration of the workspace. Also introduced is a novel end-to-end CTR generalized DeepRL policy, which to our knowledge, is the first work in generalizing over tube parameters with a model-free framework for CTRs. Thus far, in deep learning approaches for CTRs, training occurs with data from a single selected CTR system, in simulation, hardware, or a combination. However, considering that hardware systems often differ from simulation due to manufacturing inconsistencies, and the large design space of CTRs, this type of training overfits the single CTR system. To demonstrate generalization to multiple CTR systems, the work investigates the scalability and accuracy of generalization over $4$ CTR systems of varying workspace sizes. In \S \ref{sec:exp_validation}, results validating these improvements as well as error comparisons to previous tip-tracking and inverse kinematics methods are presented. Finally, we discuss strategies for translation to hardware including domain transfer and initial experiments for a domain transfer strategy known as domain randomization. The contribution evolved from our previous work \cite{iyengar2021} may be summarised as:
\begin{enumerate}
    \item Investigating constrained and constraint-free rotation on tube rotation.
    \item Development of an initial proof-of-concept CTR system generic policy for CTRs.
    \item Details for a pathway and initial simulation results for strategies to hardware translation.
\end{enumerate}

\section{Related Work}
\label{sec:related_work}
Over the last few years, deep learning approaches have become popular for control as well as kinematic and dynamic estimation of CTRs. The first deep learning approach was by Bergeles et al. \cite{Bergeles2015} for forward and inverse kinematics, performed in a simulation where a simple extension and rotation representation was used. However, tube rotation representation in the network resulted in ambiguity in inverse kinematics solutions. In later work by Grassman et al. \cite{grassmann2018}, by improving the joint representation, errors of $2.8\%$ of robot length were achieved, albeit in a limited region of the workspace and in simulation. More recently, work on shape estimation and shape-to-joint input estimation has been investigated \cite{Kuntz2020, Liang2021} using deep neural networks. Finally, Donat et al. \cite{Donat2020} introduced tip contact force estimation based on backbone deflection using a data-driven approach via deep direct cascade learning (DDCL). With tip error represented as a percentage of robot backbone length, more traditional optimization-based methods and inverse Jacobian methods have been found to have a tip tracking error of $3.2\%$ and $2.5\%$. State-of-the-art closed-loop control methods can achieve errors of $0.9\%$ and $0.5\%$ \cite{xu2014}. With active constraints and the use of a model-predictive closed loop controller, tip errors of $0.3-0.5\%$ \cite{khadem2020} have also been reported.

The Jacobian-based controllers which are common in literature \cite{xu2014, rucker2011} can be used in a closed-form fashion to perform path following with a CTR. A similar comparison was done in \cite{khadem2020}. Given a control input or desired change in joint values $\dot{q}_d$, the desired change in Cartesian space $\dot{x}_d$ and desired position in Cartesian space $x_d$, a positive semi-definite matrix $K_p$ and the pseudo-inverse Jacobian $J^\dagger$, we can define a control law such that

\begin{equation}
    \dot{q}_d = J^\dagger \left [ \dot{x}_d + K_p \left ( x_d - x \right ) \right ].
\end{equation}
Moreover, the pseudo-inverse can become very sensitive to singularities so a damping factor $\Lambda$ is added such that $J^\dagger=(J^T J + \Lambda^2 I)^{-1} J^T$. As shown in \cite{khadem2020}, this method does not account for joint limits resulting in failed trajectories. Although learning-based approaches have been well developed and have had success for forward kinematics, force estimation and shape estimation, inverse kinematics and control using deep learning has remained an open problem for CTRs. Given that the deep learning-based forward kinematics \cite{grassmann2018} and shape estimation \cite{Kuntz2020} report better error metrics than their physics-based model comparisons, investigating a deep learning-based approach for inverse kinematics and control could be beneficial and advantageous. To this end, we have investigated the use of DeepRL for CTRs. Our initial work \cite{iyengar2020} investigated the exploration problem for CTRs with simpler constant curvature dominant stiffness kinematics for simulation. The exploration problem stems from previous analysis for workspace characterization \cite{burgner2014} that has shown the bias associated with uniform joint sampling for CTR workspaces. Due to the constraints to extension from the actuation perspective, full extension and retraction are less likely to be sampled. Since DeepRL methods rely on the experiences collected during training episodes as determined by the agent's actions if full extension or retraction joint values are not sampled, kinematics and control in those areas of the workspace will not be accurate. To mitigate this, noise is usually added to the selected actions or policy network to explore the state space. Our initial work determined that applying separate noise to the rotation and extension joints was crucial in acquiring a policy with accurate control. 

As this work builds on our previous DeepRL literature, a brief overview of key concepts is introduced. In our last work \cite{iyengar2021} we improved our initial DeepRL approach \cite{iyengar2020} by using a more accurate geometrically exact kinematics model \cite{gilbert2016} for simulation, investigating joint representations, and applying a reward curriculum to improve sample efficiency (faster policy convergence with fewer data) for training. Two joint representations and three curricula functions were evaluated. The curricula were used to determine the goal tolerance during training steps, a novel approach for DeepRL to our best knowledge. The egocentric representation with the decay curriculum performed best overall in terms of sample efficiency and error metrics. To demonstrate policy robustness, a second noise-induced simulation was created where Gaussian noise was added to the join values as encoder noise and end-effector position as tracking noise. The training was then performed on the noise-induced and noise-free simulation. Performing evaluations on a noise-induced simulation, a slight improvement was seen in policy trained with the noise-induced simulation. The main takeaway from these experimental results was that the policy learned can incorporate some amount of noise in the state, and still perform adequately.

In \S \ref{sec:methods}, we first formulate the Markov Decision Process (MDP) consisting of state, action, and rewards. Next in subsection \ref{subsec:sim_env}, we review the results of our previous work, specifically joint representations of egocentric and proprioceptive, reward curricula, and combinations of representations and curricula. We then expand to constraint-free rotation in \ref{subsec:improvements} to significantly improve error metrics and finally introduce and evaluate the system identifier to generalize results across multiple CTR systems. Finally, in \S \ref{sec:exp_validation} we provide an initial experiment for domain transfer to hardware using domain randomization in simulation to motivate transfer to hardware.

\section{Methods}
\label{sec:methods}
In RL, Markov Decision Processes (MDPs) define mathematically the agent's task. Importantly, it defines the key elements for changing the state, the associated rewards, and the actions that affect the state to achieve the task.
\subsection{Markov Decision Process Formulation}
\begin{figure}[tb]
      \centering
      \includegraphics[width=0.6\linewidth]{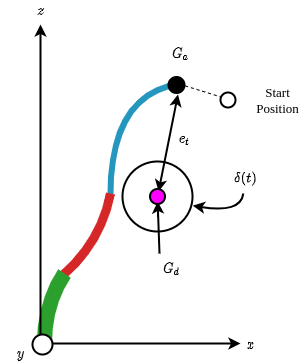}
      \caption{Illustration of the CTR system with different components of the MDP state definition including the starting position (white), achieved goal ($G_a$) (black), desired goal ($G_d$) (magenta), goal tolerance ( $\delta(t)$). The outer tube (green), middle tube (red), and inner tube (blue).}
      \label{fig:goal_diagram}
\end{figure}

In the following, the state, action, reward, and goals are defined.
\begin{figure}[tb]
  \centering
  \includegraphics[width=\linewidth]{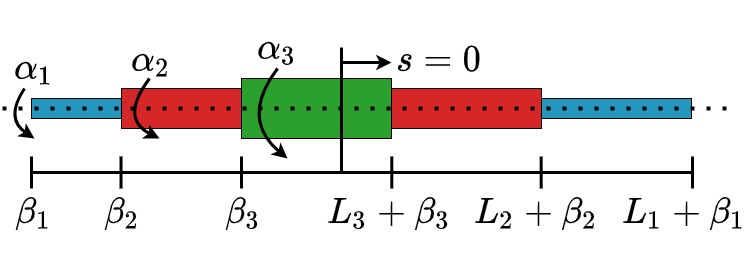}
  \caption{Extension $\beta_i$ and rotation $\alpha_i$ for tube $i$ of a $i=3$ 3-tube tube CTR system where $L_i$ is the overall length. $s$ is the arc length or axis along the backbone.}
  \label{fig:joint_values}
\end{figure}
\newline
State ($s_t$): The state at timestep $t$, is defined as the concatenation of the trigonometric joint representation \cite{grassmann2018}, Euclidean norm between the current desired position and desired position and current goal tolerance. As shown in Fig. \ref{fig:joint_values}, rotation and extension of tube $i$ (ordered innermost to outermost) are $\alpha_i$ and $\beta_i$ with $L_i$ representing the full length. First, the trigonometric representation, $\gamma_i$, of tube $i$ is defined as:
\begin{equation}
    \gamma_i = \{ \gamma_{1,i}, \gamma_{2,i}, \gamma_{3,i} \} = \{ \cos(\alpha_i), \sin(\alpha_i), \beta_i \}.
\end{equation}
The rotation can be retrieved by taking the arc-tangent
\begin{equation}
    \alpha_i =  \arctantwo (\gamma_{2,i}, \gamma_{1,i}).
\end{equation}
The extension joint $\beta_i$ can be retrieved directly and has constraints
\begin{equation}
    0 \geq \beta_3 \geq \beta_2 \geq \beta_1
\end{equation}
\begin{equation}
    0 \leq  L_3 + \beta_3 \leq L_2 + \beta_2 \leq L_1 + \beta_1
\end{equation}
from the actuation side. In our previous work, the rotation was constrained from $[-180^{\circ}, 180^{\circ}]$, which was not required in the trigonometric representation, as will be shown with constraint-free rotation. The Cartesian goal error is the current error of the achieved end-effector position $G_{a}$, and desired end-effector position $G_{d}$. Lastly, the current goal tolerance, $\delta(t)$ is included in the state where $t$ is the current timestep $t$ of training. The full state, $s_t$, can then be defined as:
\begin{equation}
    s_t = \{ \gamma_1, \gamma_2, \gamma_3, G_{a} - G_{d}, \delta(t) \}.
    \label{eqn:state}
\end{equation}
\newline
Action ($a_t$): Actions are defined as a change in rotation and extension joint positions: 
\begin{equation}
    a_t = \{ \Delta \beta_1, \Delta \beta_2, \Delta \beta_3, \Delta \alpha_1, \Delta \alpha_2, \Delta \alpha_3 \}.
\end{equation}
The maximum actions in extension and rotation are set to $1.0$ mm and $5^{\circ}$. 
\newline
Goals ($G_a$, $G_d$): Goals are defined as Cartesian points within the workspace of the robot. There is the achieved goal, $G_{a}$, and desired goal, $G_{d}$ where the achieved goal is determined with the forward kinematics of the kinematics model used and is recomputed at each timestep as the joint configuration changes from the selected actions from the policy. The desired goal updates at the start of every episode where a desired goal is found by sampling valid joint configurations in the workspace and applying forward kinematics of the model. However, this is not uniform in Cartesian space and requires action exploration.
\newline
Rewards ($r_t$): The reward is a scalar value returned by the environment as feedback for the chosen action by the agent at the current timestep. In this work, sparse rewards are used as they have been shown to be more effective than dense rewards when using hindsight experience replay (HER) \cite{andrychowicz2017}. The reward function used in this work is defined as:
\begin{equation}
    r_{t} =
        \begin{cases}
          0 & \text{if } e_t \leq \delta (t)\\
         -1 & \text{otherwise}
    \end{cases}
    \end{equation}
where $e_t$ is the Euclidean distance $||G_{a} - G_{d}||$ at timestep $t$ and $\delta(t)$ is the goal-based curriculum function that determines the goal tolerance at training timestep $t$. The workspace and various state and reward elements are illustrated in Fig. \ref{fig:goal_diagram}.
\subsection{Goal-Based Curriculum}
In our previous work, we introduced a novel goal-based curriculum that reduces goal tolerance through training steps to improve error convergence and overall error convergence. Linear and exponential decay goal-based curriculum along with a baseline constant curriculum function were compared with combinations of proprioceptive and egocentric joint representations. Each curriculum reduces the goal tolerance as a function of timestep $t$, the number of timesteps to apply the function, $N_{ts}$, the initial tolerance, $\delta_{initial}$ and final tolerance, $\delta_{final}$. The linear curriculum is defined as
\begin{equation}
\begin{aligned}
    \delta_{linear} (t) &= at + b \\
    a &= \frac{\delta_{final} - \delta_{initial}}{N_{ts}} \\
    b &= \delta_{initial},
\end{aligned}
\label{eqn:linear}
\end{equation}
and the exponential decay curriculum is defined as
\begin{equation}
\begin{aligned}
    \delta_{decay} (t) &= a (1-r)^t \\
    a &= \delta_{initial} \\
    r &= 1 - \left( \frac{\delta_{final}}{\delta_{initial}} \right) ^{\frac{1}{N_{ts}}}.
\end{aligned}
\label{eqn:decay}
\end{equation}
The values used for these various parameters can be found in \cite{iyengar2021}.

\subsection{Joint Representation}
\begin{figure}[tb]
     \subfloat[]{\includegraphics[width=0.24\textwidth]{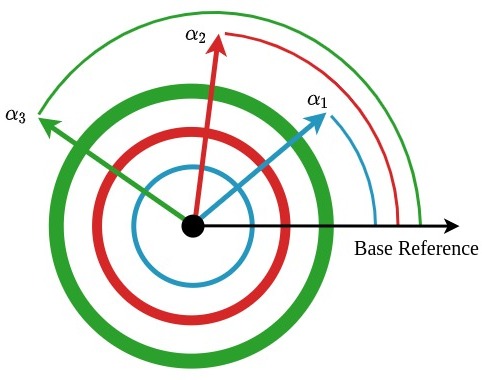}\label{fig:proprioceptive}}%
     \subfloat[]{\includegraphics[width=0.24\textwidth]{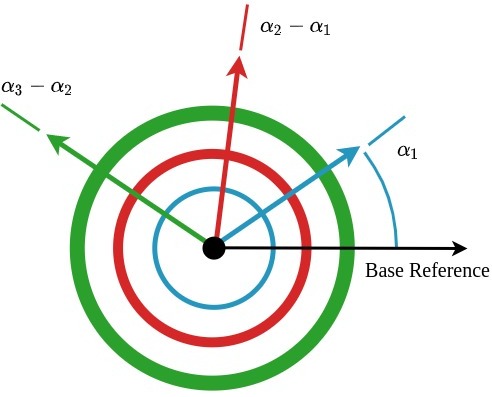}\label{fig:egocentric}}%
     \caption{Illustration of rotational \protect\subref{fig:proprioceptive} proprioceptive and \protect\subref{fig:egocentric} egocentric joint representation for rotational joints. Inner (blue), middle (red), and outer (green) tubes' rotation representation with respect to the base (black).}
    \label{fig:joint_representations}
\end{figure}

To improve learning sample efficiency, joint representations were investigated. Specifically, proprioceptive (absolute) and egocentric (relative) joint representations where the reference of measure for each joint position is changed respectively. Although proprioceptive representations are used often in classical control for robotics, egocentric joint representations are utilized heavily in reinforcement learning control simulation environments like the DeepMind control suite \cite{dmcontrolsuite2020}. In proprioceptive or absolute joint representation all the joints are referenced from a common base reference. This is illustrated for rotations in Fig. \ref{fig:proprioceptive}. However, in egocentric or relative joint representations, only the inner tube is referenced from the base shown in Fig. \ref{fig:egocentric}. The next outer tube is referenced from the previous inner tube and so forth. This can be used for both rotation and extension joints.
\begin{equation}
\begin{aligned}
\alpha_{ego} &= \{ \alpha_1, \alpha_{2}-\alpha_{1},\alpha_{3}-\alpha_{2}\} \\
&= \{ \alpha_1, \Delta \alpha_{2-1},\Delta \alpha_{3-2}\}
\end{aligned}
\end{equation}
and extensions
\begin{equation}
\beta_{ego} = \{ \beta_1, \beta_{2}-\beta_{1}, \beta_{3}-\beta_{2}\}
\end{equation}
To retrieve the absolute joint representation, the cumulative sum is taken as shown below: 
\begin{equation}
    \alpha_{prop} = \{ \alpha_1, \Delta \alpha_{2-1} + \alpha_1,  \Delta \alpha_{3-2} + \Delta \alpha_{2-1} + \alpha_1 \}
\end{equation}

\subsection{CTR Simulation Environment}\label{subsec:sim_env}
To collect data and experiences for the deepRL algorithm to learn a policy, a simulation environment for kinematics was developed following the openAI gym framework \cite{brockman2016}. With this gym framework, any compatible DeepRL algorithm can be utilized to learn a policy for the given task. The environment takes a set of tube parameters describing a CTR system, joint configuration, and selected actions by the agent to determine the overall shape of the CTR. For DeepRL, a large number of experiences are needed to train a policy, so a "sweet-spot" or relatively fast computationally and relatively accurate kinematics model was used which was first presented in \cite{webster2008} and later presented for externally loaded CTRs with point and distributed forces and moments in \cite{rucker2010, dupont2012}. This model ignores friction, permanent strain, and forces along the backbone of the robot. The gym framework is then used for experiments including inverse kinematics, path following, and domain randomization.

To summarize the results of our previous work, investigating combinations of reward curriculum and joint representations, the egocentric, decay curriculum was found to perform best in evaluation. Training was done with deep deterministic policy gradient (DDPG) \cite{lillicrap2015} algorithm with hindsight experience replay (HER) \cite{andrychowicz2017}. Evaluation was done with $1000$ evaluation episodes with results shown in Table \ref{tab:icra_results}. The success rate is defined as the number of successful trajectories over the total number of trajectories, where success is achieving a trajectory with below $1$ mm error. In the simulation environment for these experiments, tube rotations were constrained from $-\pi$ to $\pi$. Unconstraining the tube rotation will be investigated in the next section. With learned policy, since the desired goal is within the state as shown in equation (\ref{eqn:state}), a policy controller was created where the desired goal was updated via a trajectory generator. This type of controller is normally not available for other deep neural network approaches as they are not inherently timestep based whereas DeepRL aims to optimize actions at each timestep. Following various paths including circular and straight-line, the mean tracking error was $0.58$ mm using the egocentric decay curriculum. To demonstrate robustness, tracking and joint noise of $0.8$ mm and $1^\circ$ was induced into the simulation resulting in $1.37$ mm mean tracking error. However, with the tube rotation constraints, exploration may not be exploring the workspace in its entirety. Another factor is workspace size which depends on the CTR system used for training. It is evident larger workspaces will take longer computation time to train. First, performing a workspace and joint error analysis, we found removing constraints on tube rotations provided a significant performance improvement, particularly in larger CTR systems. Second, we found generalization possible with the MDP formulation by appending a system identifier to the state. This generic policy would be useful as only a single policy would be needed for multiple systems and would be the first step towards full generalization for deep learning-based CTR kinematics. Moreover, this generalization motivates experiments using domain randomization, for domain transfer from simulation to hardware. In the following, we describe the methods for the two main experiments, improvement through constraint-free rotation and generalization.

\begin{table}[tb]
\centering
\caption{Joint representation with curriculum results.}
\label{tab:icra_results}
\begin{tabular}{@{}cccc@{}}
\toprule
\textbf{\begin{tabular}[c]{@{}c@{}}Joint\\ Representation\end{tabular}} & \textbf{Curriculum} & \textbf{\begin{tabular}[c]{@{}c@{}}Error\\ (mm)\end{tabular}} & \textbf{\begin{tabular}[c]{@{}c@{}}Success\\ rate\end{tabular}} \\ \midrule
\multirow{3}{*}{\textbf{Proprioceptive}}                                & Constant            & 2.43 $\pm$ 0.09                                               & 0.85                                                            \\ \cmidrule(l){2-4} 
                                                                        & Linear              & $3.31 \pm 0.15$                                               & 0.87                                                            \\ \cmidrule(l){2-4} 
                                                                        & Decay               & $3.16 \pm 0.12$                                               & 0.80                                                            \\ \midrule
\multirow{3}{*}{\textbf{Egocentric}}                                    & Constant            & $1.94 \pm 0.05$                                               & 0.87                                                            \\ \cmidrule(l){2-4} 
                                                                        & Linear              & $3.38 \pm 0.15$                                               & 0.89                                                            \\ \cmidrule(l){2-4} 
                                                                        & Decay               & $1.29 \pm 0.03$                                               & 0.93                                                            \\ \bottomrule
\end{tabular}
\end{table}

\subsection{Improvements with constraint-free tube rotation}\label{subsec:improvements}
\begin{figure*}[tb]
\subfloat[\label{fig:const_ag_errors}]{%
  \includegraphics[width=0.5\linewidth]{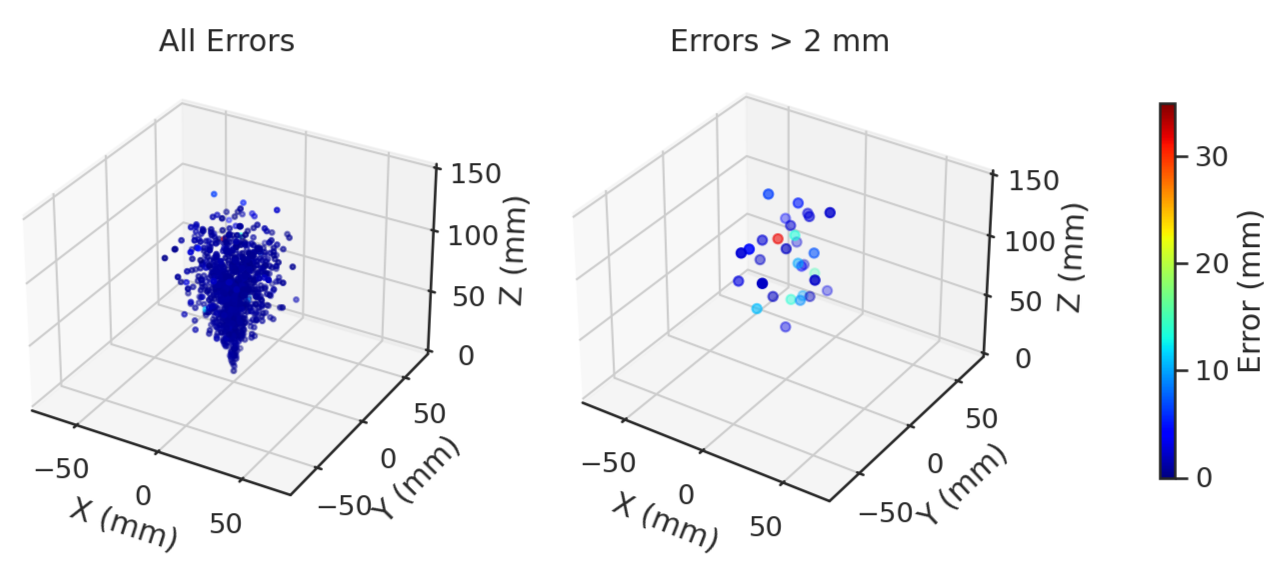}%
}\hfill
\subfloat[\label{fig:free_ag_errors}]{%
  \includegraphics[width=0.45\linewidth]{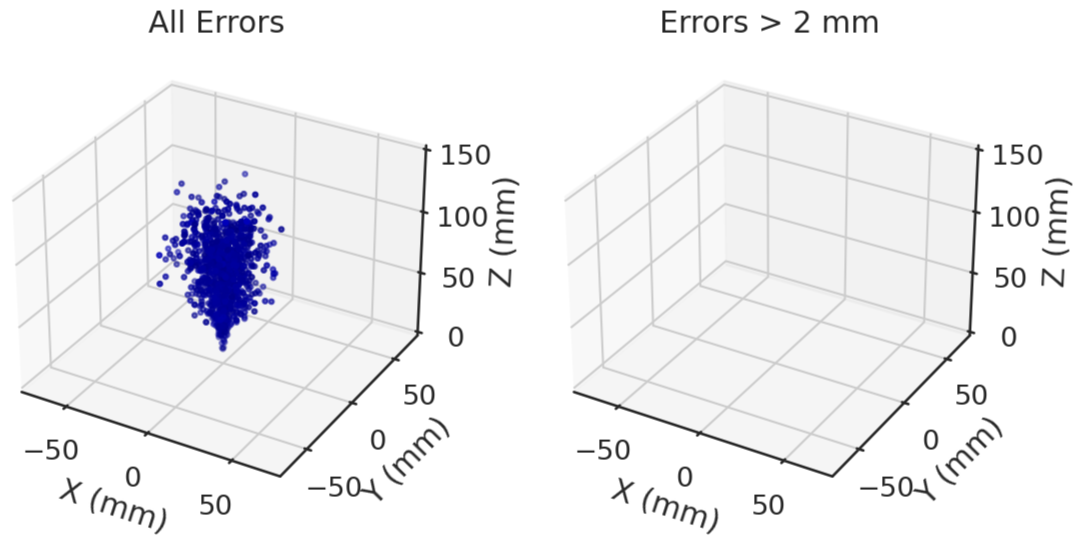}%
}
\caption{Plotting achieved goal positions of evaluations of a policy trained in a simulation environment with constrained tube rotation \protect\subref{fig:const_ag_errors} and \protect\subref{fig:free_ag_errors} constraint-free tube rotations. Furthermore, each plot is filtered to errors greater than $2$ mm to the desired goal of the episode to illustrate large errors. In \protect\subref{fig:free_ag_errors}, no errors are greater than $2$ mm demonstrating a low error variance.}
\label{fig:const_unconst_workspace}
\end{figure*}

With the best policy training method, egocentric decay from the previous work, state information such as the achieved goal, desired goal, Cartesian error, and joint error at the end of each episode from $1000$ evaluation episodes were tabulated after training the policy on the larger CTR system $0$. Plotting the Cartesian points of the achieved goal with RGB values corresponding to Cartesian error to the desired goal results in Fig. \ref{fig:const_ag_errors}. Furthermore, thresholding achieved goal points with Cartesian errors to the desired goal greater than $2$ mm, and regions of larger errors can be isolated. As seen, there is a large standard deviation in errors greater than $2$ mm with constrained rotation. This motivates unconstraining tube rotations during DeepRL training for exploration. Shown in Figure \ref{fig:free_ag_errors} are constraint-free rotation training results in no errors greater than $2$ mm, thus the standard deviation of errors is greatly reduced.

\begin{figure}[tb]
      \centering
      \includegraphics[width=0.8\linewidth]{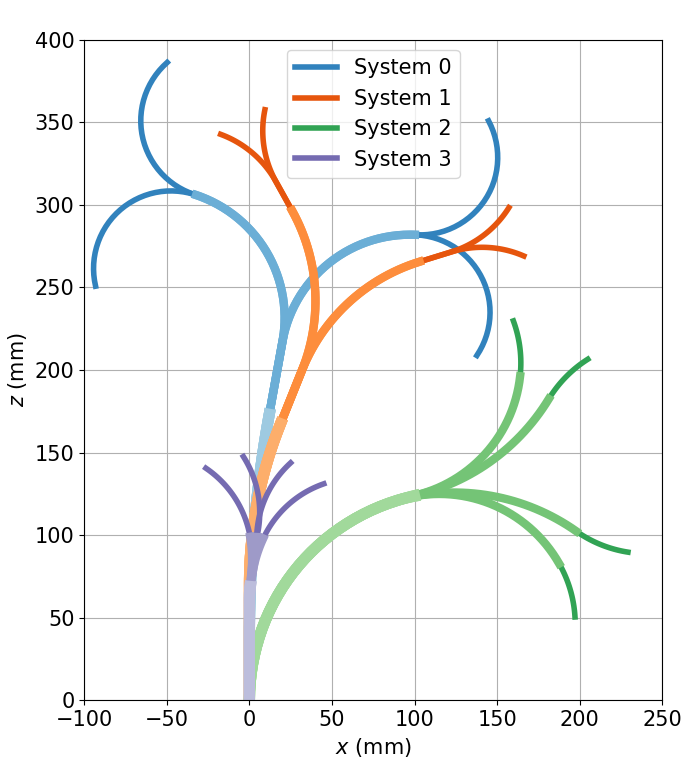}
      \caption{Visualizing the different CTR systems used with the geometrically exact modeling. Chosen were $4$ different CTR systems, each composed of $3$ tubes of various parameters resulting in different workspaces. The longest system is CTR system 0 (blue), with the shortest CTR system being system $3$ (purple). System $2$ (green) has the largest curvature resulting in a wide workspace. Finally, the CTR system $1$ is in orange. The systems are ordered from longest to shortest. Each system in the figure is fully extended with the middle and inner tube rotated at $0^\circ$ and $180^\circ$ for a total of $4$ configurations per system.}
      \label{fig:ctr_systems}
\end{figure}

\begin{figure*}[tb]
\centering
\subfloat[\label{fig:const_ag_joint_errors_1}]{\includegraphics[width=0.3\textwidth]{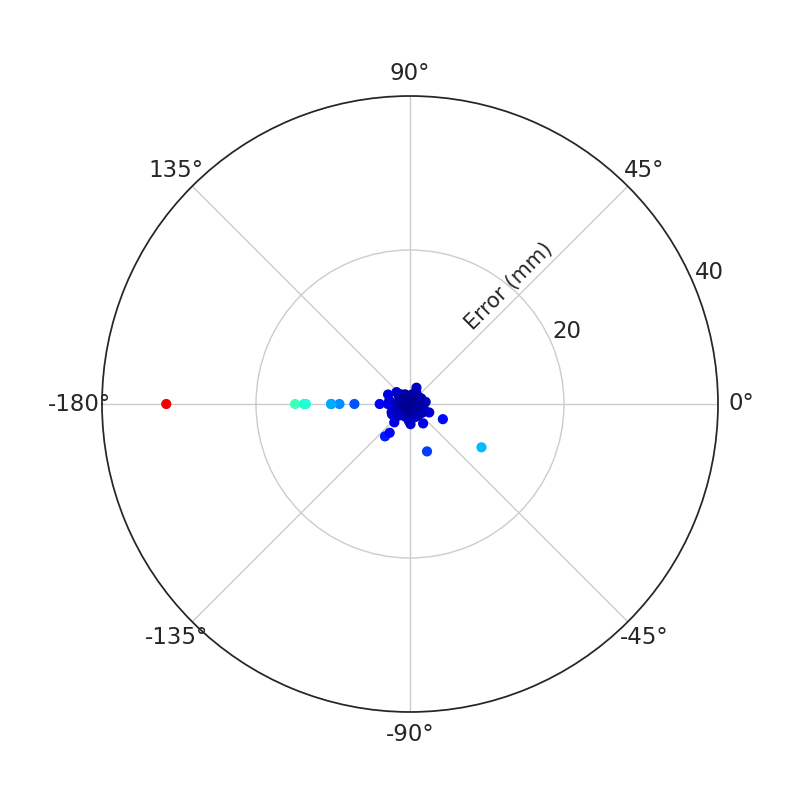}}\hfill
\subfloat[\label{fig:const_ag_joint_errors_2}]{\includegraphics[width=0.3\textwidth]{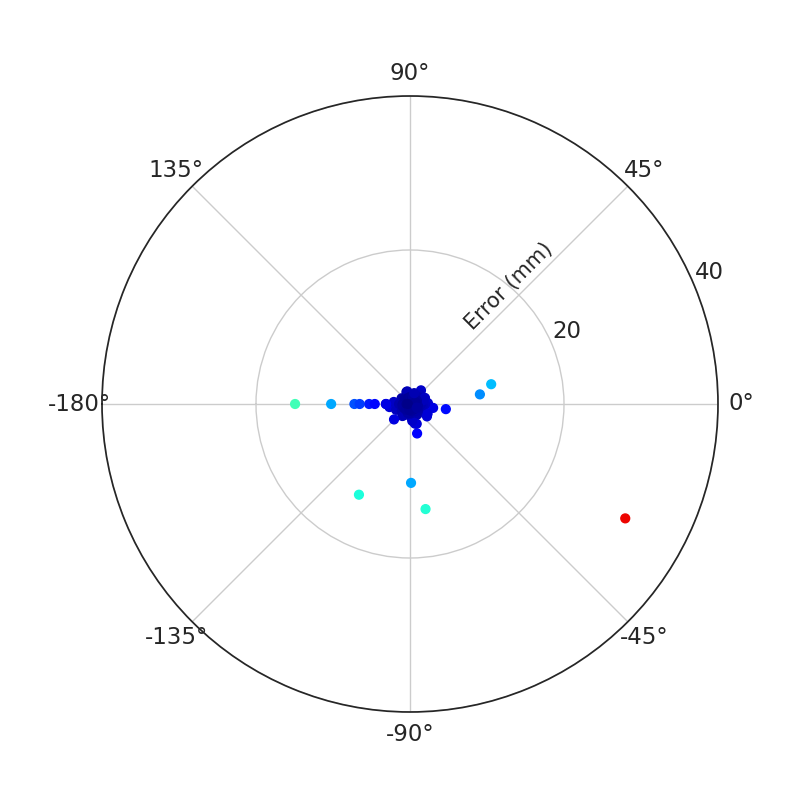}}\hfill
\subfloat[\label{fig:const_ag_joint_errors_3}]{\includegraphics[width=0.3\textwidth]{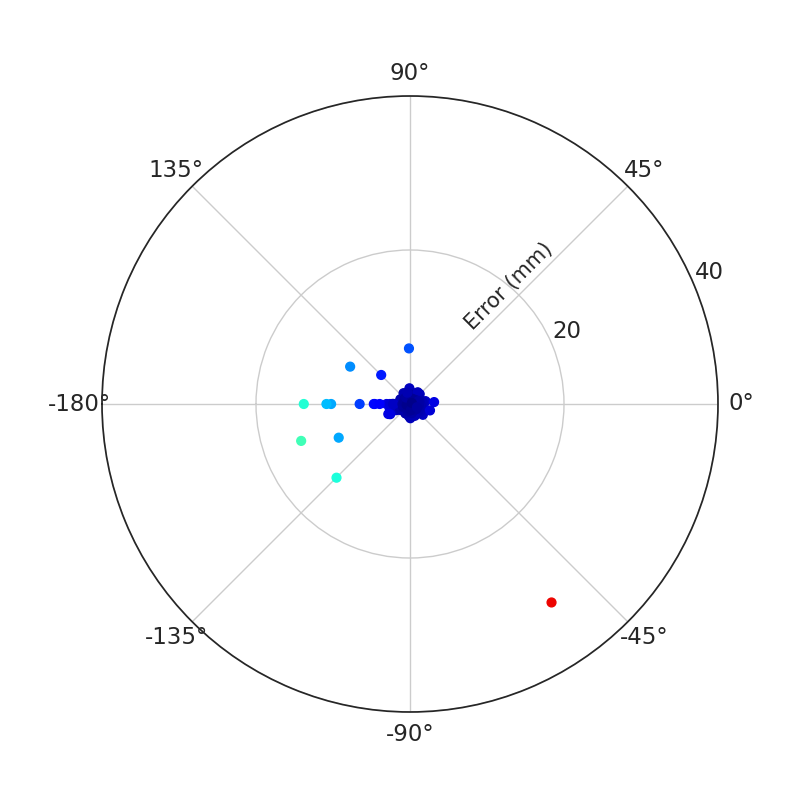}}\\
\subfloat[\label{fig:free_ag_joint_errors_1}]{\includegraphics[width=0.3\textwidth]{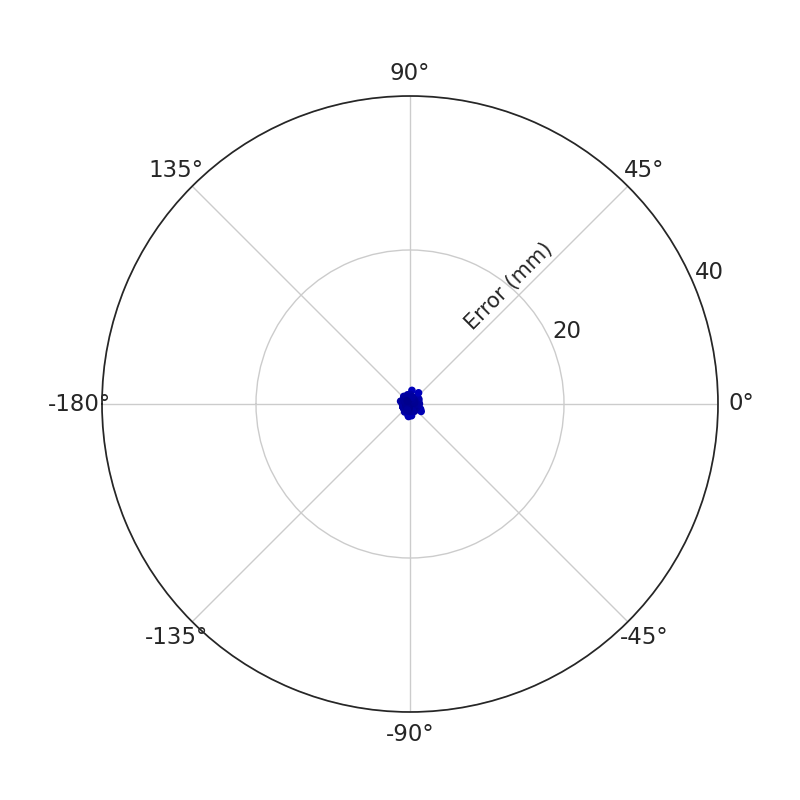}}\hfill
\subfloat[\label{fig:free_ag_joint_errors_2}]{\includegraphics[width=0.3\textwidth]{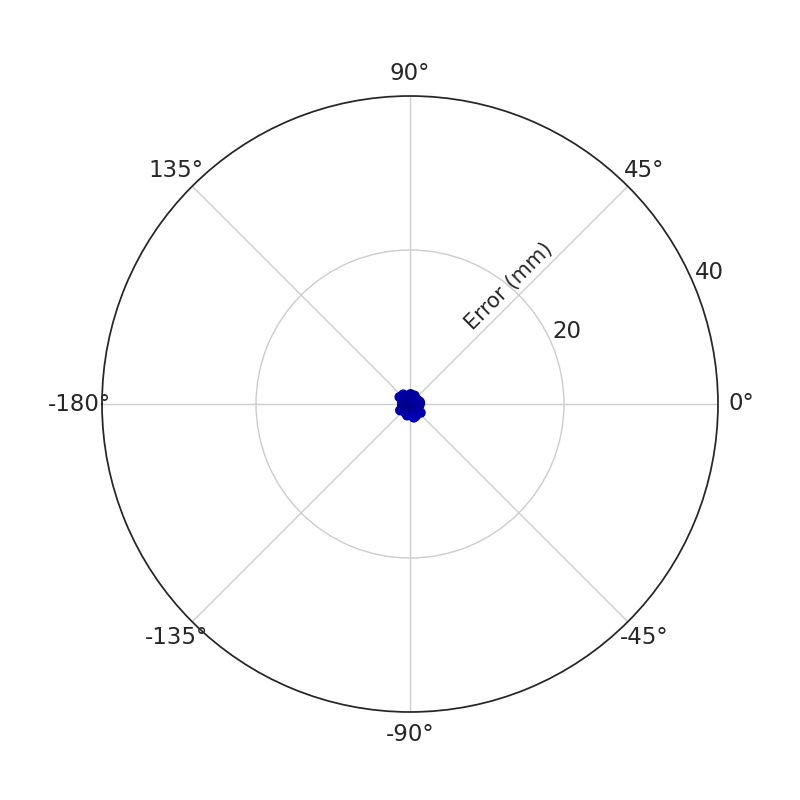}}\hfill
\subfloat[\label{fig:free_ag_joint_errors_3}]{\includegraphics[width=0.3\textwidth]{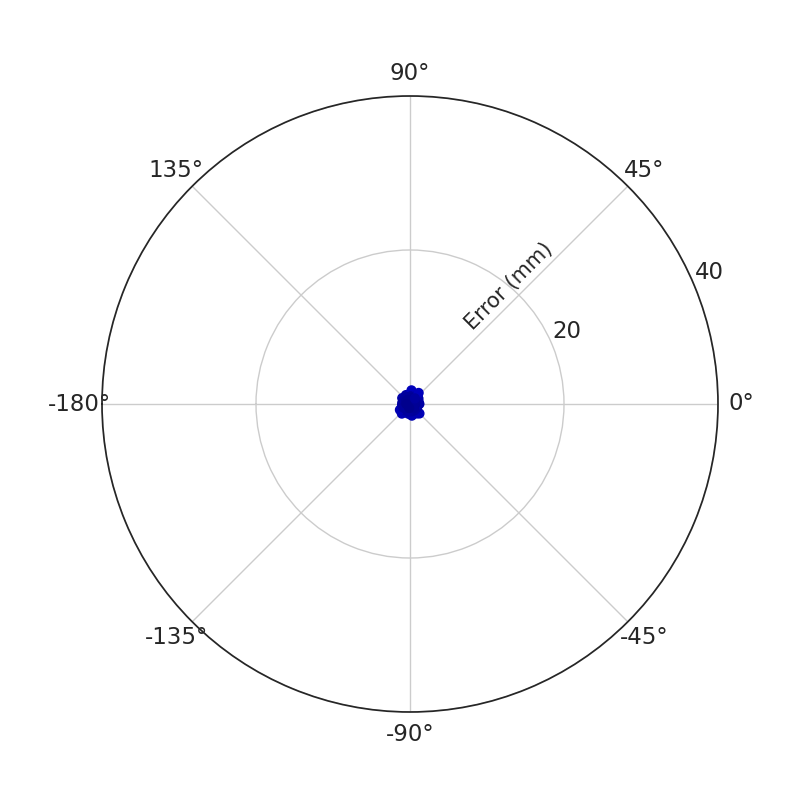}}
\caption{From evaluations, plotting tube rotations and associated errors in RGB. Constrained rotation errors with respect to tube rotation angle for \protect\subref{fig:const_ag_joint_errors_1} $\alpha_1$, \protect\subref{fig:const_ag_joint_errors_2} $\alpha_2$ and \protect\subref{fig:const_ag_joint_errors_3} $\alpha_3$. Constraint-free rotation errors with respect to tube rotation angle for \protect\subref{fig:free_ag_joint_errors_1} $\alpha_1$, \protect\subref{fig:free_ag_joint_errors_2} $\alpha_2$ and \protect\subref{fig:free_ag_joint_errors_3} $\alpha_3$. See Fig. \ref{fig:const_unconst_workspace} for RGB color legend.\label{fig:polar_plots}}
\end{figure*}

In order to investigate the joint values associated with these large errors, Figure \ref{fig:const_unconst_workspace} shows that the Cartesian achieved goals for the innermost tube or $\alpha_1$ at the end of each episode and the associated errors in the robot workspace. As seen, there is a large number of errors greater than $2$ mm with some points up to $30$ mm in error. In Figs. \ref{fig:const_ag_joint_errors_1}, \ref{fig:const_ag_joint_errors_2} and \ref{fig:const_ag_joint_errors_3}, the constraints causing the large errors at the boundaries of $-180$ and $+180$ where the largest outliers for errors are confirmed with a polar plot for each tube rotation. In previous work, this rotation constraint is to limit the joint space sampling during the generation of new desired goals, starting joint values and data collection as has been implemented in previous CTR deep learning work \cite{grassmann2018}.
However, this constraint through timesteps is non-essential in the trigonometric representation. Training a policy using the egocentric decay curriculum without constraining the rotations of the tubes considerably improved the error metrics from the previously constrained egocentric curriculum from a mean error of $4.05$ mm to $0.68$ mm for the largest system $0$.

\begin{table}[tb]
\centering
\caption{Constrained and unconstrained rotation experiments for various CTR systems.}
\label{tab:rotation_exps}
\begin{tabular}{@{}cccc@{}}
\toprule
\textbf{Experiment}                                                                 & \textbf{System} & \textbf{Error (mm)} & \textbf{Success rate} \\ \midrule
\multirow{4}{*}{\textbf{Constrained}}                                               & 0               & $4.05 \pm 18.18$    & $77.1$                \\ \cmidrule(l){2-4} 
                                                                                    & 1               & $2.53 \pm 12.40$    & $92.5$                \\ \cmidrule(l){2-4} 
                                                                                    & 2               & $3.13 \pm 16.08$    & $92.8$                \\ \cmidrule(l){2-4} 
                                                                                    & 3               & $0.52 \pm 1.49$     & $99.6$                \\ \midrule
\multirow{4}{*}{\textbf{\begin{tabular}[c]{@{}c@{}}Constraint\\ free\end{tabular}}} & 0               & $0.68 \pm 0.28$     & $98.3$                \\ \cmidrule(l){2-4} 
                                                                                    & 1               & $0.64 \pm 0.67$     & $99.3$                \\ \cmidrule(l){2-4} 
                                                                                    & 2               & $0.61 \pm 0.24$     & $99.0$                \\ \cmidrule(l){2-4} 
                                                                                    & 3               & $0.45 \pm 0.20$     & $99.9$                \\ \bottomrule
\end{tabular}
\end{table}
\begin{figure}[tb]
\subfloat[\label{fig:tro_free_0_ik_1}]{%
  \includegraphics[width=0.5\linewidth]{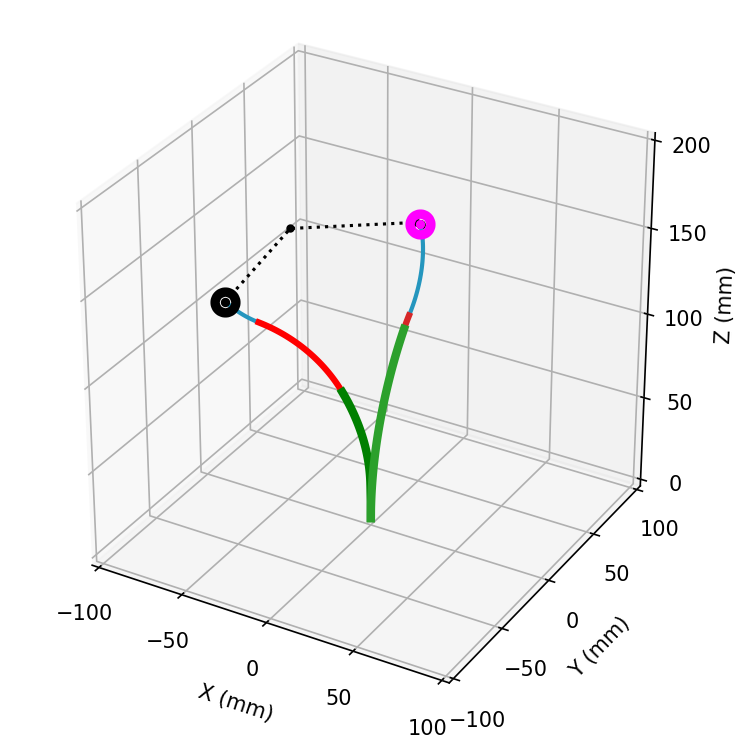}%
}\hfill
\subfloat[\label{fig:tro_free_0_ik_2}]{%
  \includegraphics[width=0.5\linewidth]{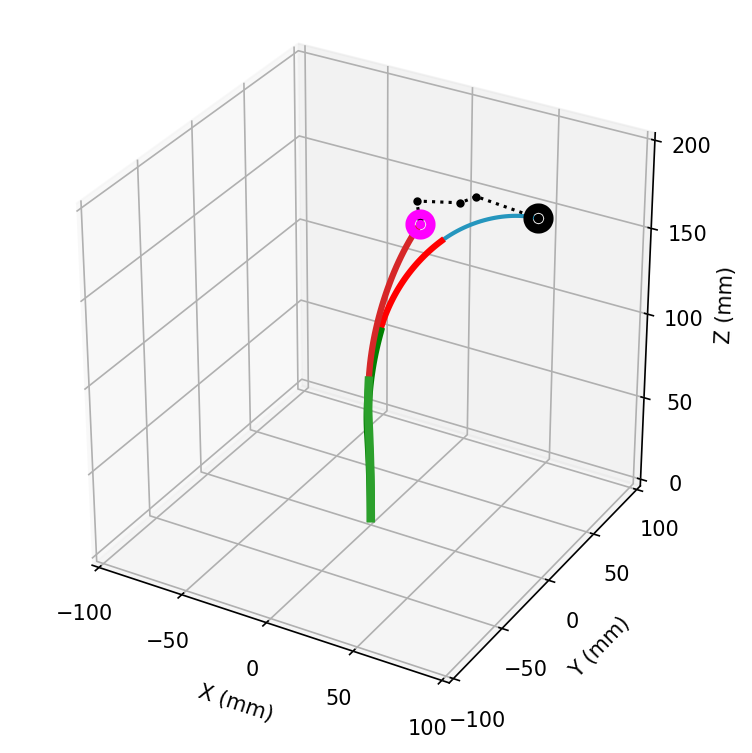}%
}
\caption{Two inverse kinematics solutions with different starting joint configurations given the same desired goal. The black dot indicates starting position and the magenta dot indicates the desired position with the smaller black dots indicating intermediary reached points.}
\label{fig:tro_ik_sol}
\end{figure}

To further analyze the behavior of the agent with respect to error, a goal distance to Cartesian error analysis was performed from the $1000$ evaluation episodes, and state information was tabulated. This analysis reveals the relationship between distance to the desired goal and the associated final error. Generally, a closer goal, ie. a low initial goal distance, would be expected to have smaller errors overall, with farther goals having larger errors. For this analysis, all $1000$ data points were used to determine the linear relationship between the initial goal distance and final Cartesian error. For constrained rotation with system $0$, the slope was found to be $3.27$ mm with a $y$-intercept of $0.8$ mm. This suggests a poor inverse kinematics solver as errors become very large with higher goal distances. With constraint-free rotation, the slope was found to be $0.6$ mm with a $y$-intercept of $0.66$ mm, a reasonable slope for such a relationship.

As an inverse kinematics solver, our DeepRL method performs adequately when rotations are unconstrained. However, an important note is only one solution is provided as it is an iterative solver and is dependent on the initial condition of the joint configuration at the start of each episode. However, this can be remedied with multiple episodes with different initial joint configurations. Visualized in Fig. \ref{fig:tro_ik_sol} is an example of the same desired end-effector position with two different initial joint configurations, resulting in two different final inverse kinematics solutions. Using the constraint-free egocentric decay method for system $0$, the desired goal position was $\left (0, 50, 150 \right )$ mm. The final joint configuration in Fig. \ref{fig:tro_free_0_ik_1} was $\beta = [-15.35, -12.59, -4.01]$ mm, $\alpha = [-3.73, -183.07, -25.54]^\circ$ with a tip error of $0.99$ mm and for Figure \ref{fig:tro_free_0_ik_2} $\beta = [-14.77, -9.12, -5.12]$ mm and $\alpha = [-179.19, -180.32, 1.93]^\circ$ with a tip error of $0.36$ mm. 

To verify the constraint-free results and to demonstrate training and evaluation of our DeepRL method, we apply it to three other CTR systems from various sources in the literature and trained each with constrained and constraint-free rotation, and performed $1000$ evaluation episodes. The main changes to hyperparameters were $3$ million training timesteps with $1.5$ million steps for the curriculum, a buffer size of $500,000$, and neural network layer size of $3$ hidden units with $256$ neurons each. The results for all four CTR systems are presented in Table \ref{tab:rotation_exps}. Of note is the increased standard deviation in constrained rotation. However, with smaller systems such as system $3$, this is not as pronounced due to the smaller robot workspace. From our previous work, there has been a large reduction in mean and standard deviation of errors as shown in the largest CTR system.

\begin{figure*}[tb]
      \centering
      \includegraphics[width=\linewidth]{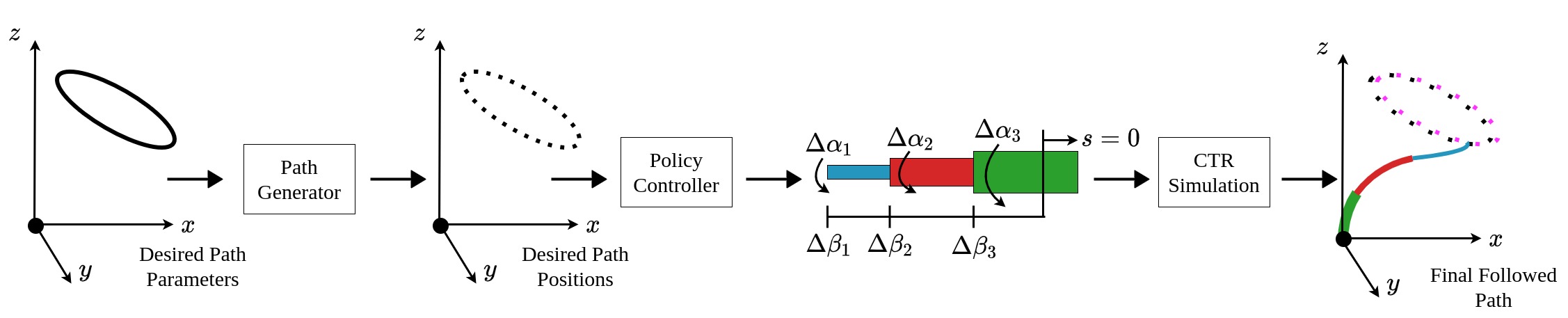}
      \caption{Illustration of the process by which paths are generated, control actions are determined, and finally, paths followed.}
      \label{fig:path_following}
\end{figure*}
\subsection{Generic Policy}
To further motivate the utility of DeepRL for CTRs, we introduce an initial proof of concept for a CTR system generic policy. A hurdle currently with using deep learning approaches for CTRs is the limitation of CTR system generalization. Because deep learning relies solely on the data collected, if only one CTR system is used for training, the learned policy will accurately control that system alone. Moreover, we aim to demonstrate that using our egocentric decay goal-based curriculum, with constraint-free rotation has improved error metrics when compared with the one that does not employ these extensions for such a generic policy.

This CTR generic policy will seek to generalize over four CTR systems that have different tube parameters. The objective is to obtain good performance across the CTR systems with a single control policy. For generalization, a system specifier, $\psi = \{0, 1, 2, 3\}$, was appended to the state, $s_t$, for the agent to differentiate the CTR systems. The state is now defined as
\begin{equation}
    s_t = \{ \gamma_1, \gamma_2, \gamma_3, G_{a} - G_{d}, \delta(t), \psi \}
    \label{eqn:new_state}
\end{equation}

\begin{table*}[t]
\centering
\vspace{5pt}
\caption{Error metrics for system generalization for the various number of systems and combination of systems. Errors presented in millimeters mean $\pm$ standard deviation.}
\label{tab:generic_system_exps}
\begin{tabular}{@{}cccccc@{}}
\toprule
\textbf{Number of Systems}              & \textbf{Systems} & \textbf{First System} & \textbf{Second System} & \textbf{Third System} & \textbf{Fourth System} \\ \midrule
\multirow{6}{*}{\textbf{Two Systems}}   & {[}0,1{]}        & $0.715 \pm 0.057$     & $0.710 \pm 0.054$      & --                    & --                     \\ \cmidrule(l){2-6} 
                                        & {[}0,2{]}        & $0.663 \pm 0.069$     & $0.650 \pm 0.063$      & --                    & --                     \\ \cmidrule(l){2-6} 
                                        & {[}0,3{]}        & $0.660 \pm 0.056$     & $0.593 \pm 0.052$      & --                    & --                     \\ \cmidrule(l){2-6} 
                                        & {[}1,2{]}        & $0.606 \pm 0.056$     & $0.623 \pm 0.064$      & --                    & --                     \\ \cmidrule(l){2-6} 
                                        & {[}2,3{]}        & $0.655 \pm 0.057$     & $0.602 \pm 0.046$      & --                    & --                     \\ \cmidrule(l){2-6} 
                                        & {[}1,3{]}        & $0.636 \pm 0.0533$    & $0.582 \pm 0.052$      & --                    & --                     \\ \midrule
\multirow{4}{*}{\textbf{Three Systems}} & {[}0,1,2{]}      & $0.726 \pm 0.092$     & $0.690 \pm 0.057$      & $0.749 \pm 0.087$     & --                     \\ \cmidrule(l){2-6} 
                                        & {[}0,1,3{]}      & $0.826 \pm 0.102$     & $0.764 \pm 0.081$      & $0.713 \pm 0.056$     & --                     \\ \cmidrule(l){2-6} 
                                        & {[}0,2,3{]}      & $0.901 \pm 4.170$     & $0.964 \pm 0.217$      & $0.791 \pm 0.053$     & --                     \\ \cmidrule(l){2-6} 
                                        & {[}1,2,3{]}      & $0.675 \pm 0.059$     & $0.647 \pm 0.074$      & $0.636 \pm 0.071$     & --                     \\ \midrule
\textbf{Four Systems}                   & {[}0,1,2,3{]}    & $0.754 \pm 0.083$     & $0.723 \pm 0.061$      & $0.727 \pm 0.069$     & $0.629 \pm 0.052$      \\ \bottomrule
\end{tabular}
\end{table*}

\begin{figure*}[tb]
     \subfloat[\label{fig:helix_1}]{\includegraphics[width=0.25\textwidth]{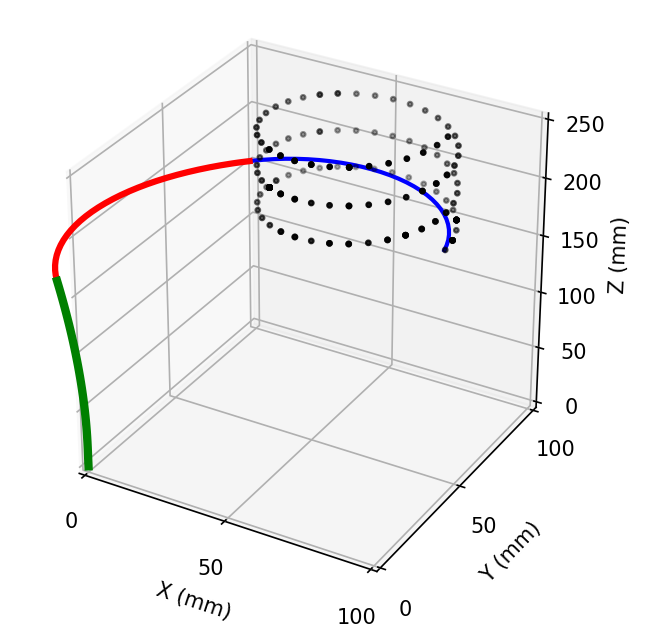}}%
     \subfloat[\label{fig:helix_2}]{\includegraphics[width=0.25\textwidth]{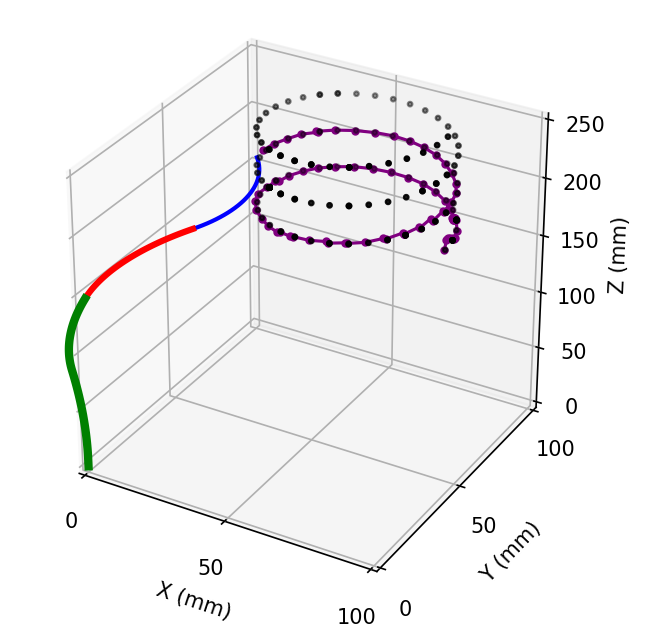}}%
     \subfloat[\label{fig:helix_3}]{\includegraphics[width=0.25\textwidth]{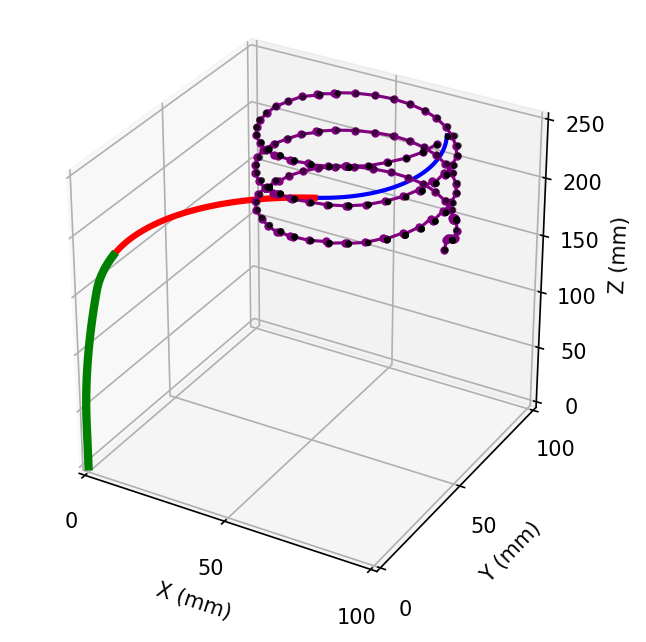}}%
     \subfloat[\label{fig:helix_full}]{\includegraphics[width=0.25\textwidth]{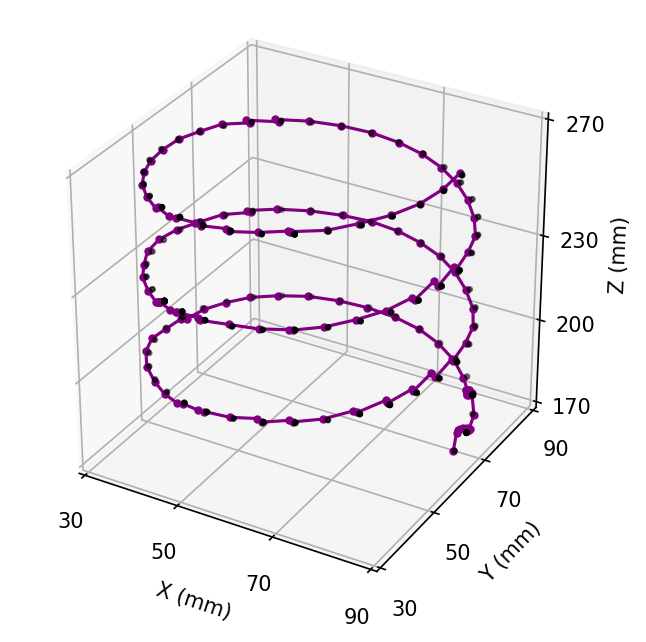}}%
     \caption{CTR system $0$ following a helix path. \protect\subref{fig:helix_1} starting point of path, \protect\subref{fig:helix_2} midway through path, \protect\subref{fig:helix_3} final point of path and \protect\subref{fig:helix_full} full path following results.\label{fig:sample_gen_helix}}
\end{figure*}
\begin{figure*}[tb]
     \subfloat[\label{fig:noisy_straight_1}]{\includegraphics[width=0.25\textwidth]{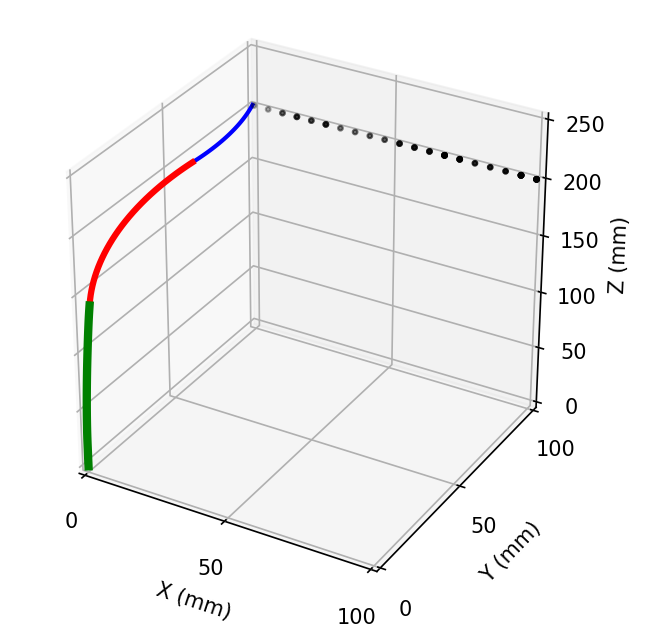}}%
     \subfloat[\label{fig:noisy_straight_2}]{\includegraphics[width=0.25\textwidth]{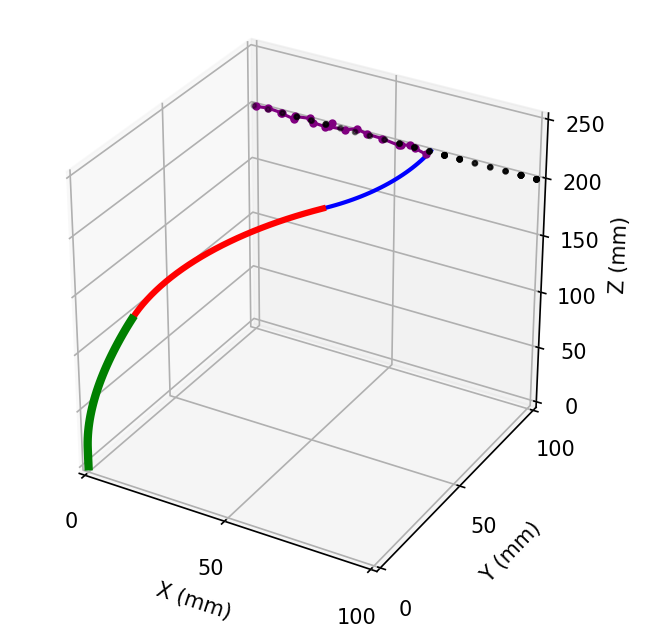}}%
     \subfloat[\label{fig:noisy_straight_3}]{\includegraphics[width=0.25\textwidth]{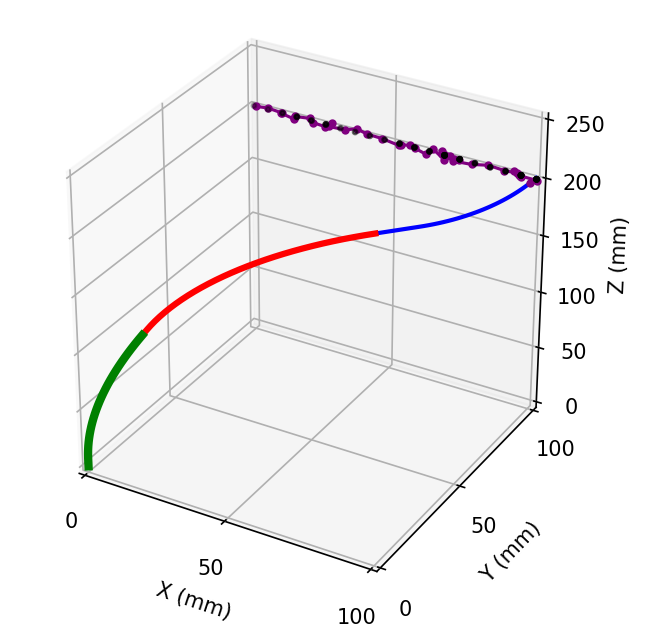}}%
     \subfloat[\label{fig:noisy_straight_full}]{\includegraphics[width=0.25\textwidth]{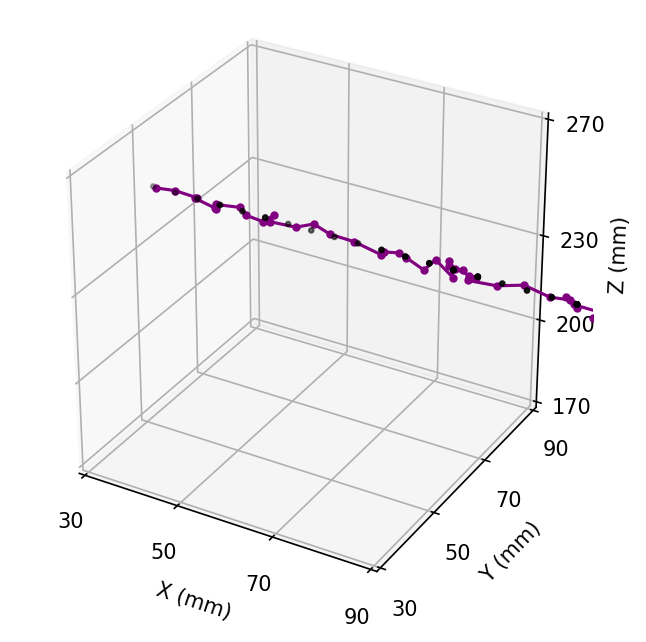}}%
     \caption{System $1$ following a straight path. \protect\subref{fig:noisy_straight_1} starting point of path, \protect\subref{fig:noisy_straight_2} midway through path, \protect\subref{fig:noisy_straight_3} final point of path and \protect\subref{fig:noisy_straight_full} full path following results.\label{fig:sample_gen_straight}}
\end{figure*}

At the start of each episode, a discrete uniform distribution is sampled to determine the value of $\psi$, or the CTR system parameters. These parameters are then set in the simulation environment and for that episode, the selected system is the one simulated for the agent's task. Once the episode is reset, a new $\psi$ is sampled. Thus, over timesteps, all systems should be sampled with the agent collecting experiences from all systems uniformly. We acknowledge this is not a true generalization, and the policy only learns the systems given, however, this initial proof-of-concept demonstrates some form of generalization is possible for CTRs using DeepRL, an attribute not shown in any previous work. Given the right network parameters, all attributes defining a CTR system could be included in the state to  generalize fully. To demonstrate initial generalization, we train a single policy to generalize over two, three, and four systems including different combinations. The systems are ordered $0$ to $3$, from the shortest overall system to the longest overall system. First, to validate our egocentric decay constraint-free method, we compared generalization with our constraint-free egocentric decay method to the constraint-free proprioceptive constant method and constrained proprioceptive constant policy. We aim to demonstrate that our constraint-free egocentric curriculum is key to policy convergence for generalization. A full set of results for all combinations of generalization are provided for the different systems. Then, to demonstrate the learned policy we present path following task results generalized over four systems, with and without sensor and encoder noise to display robustness.

We make our gym environment and the code to reproduce these results available online. \footnote{[Online]. Available: https://github.com/keshaviyengar/gym-ctr-reach} 
\section{Experiments and Results}
\label{sec:exp_validation}
Thus far, we have verified that constrained tube rotation causes large errors at the boundary of the constraints. This is due to constraints on rotation resulting in inadequate exploration. Furthermore, we verified the method as an inverse kinematics solver by solving for a desired goal position with two different initial starting positions. The result was two different inverse kinematics solutions given the two different initial starting positions. As mentioned prior, the deepRL method can also perform path following. To validate this, experiments in simulation were performed following various paths with the constraint-free egocentric decay DeepRL method. Next, we investigate the state identifier for generalization, and present error metrics and path following results.

In a surgical scenario, the surgeon would be controlling the end-effector tip position with a haptic device similar to a Phantom Omni from Sensable Technologies. Thus, the inputs are Cartesian coordinates of the end effector tip position and fit well with the state description of our DeepRL method. Therefore, there is the single-point inverse kinematics task as seen previously, and the path-following tasks described here. The experimental framework for path-following tasks is as follows. First, $x,y,z$ desired goal positions are generated by a path generator component to substitute control inputs from a user. This component takes as input path shape parameters and the discretization parameter. Shapes include polygons, circular and helix paths. The generator outputs a series of $x,y,z$ desired goal positions of the path with the path parameters given. The next component is the policy controller which takes two inputs, the desired goal positions and the initial joint configuration. The controller is described in detail in the next section. The controller outputs the changes in joint positions to achieve the path by reaching each of the desired goals given by the path generator. The controller is open-loop since information about whether the goal is reached is not relayed back. There are a number of steps given to reach the goal, and even if the goal is not reached, the next goal is given. Finally, the last component is the CTR simulation. The simulation takes an input of changes in joint positions, performs kinematics for each step, and returns the achieved goal positions of the end-effector as well as the full CTR backbone shape, with which the resulting path followed can be visualized. The entirety of the validation path following framework can be found in Fig. \ref{fig:path_following}.
\subsection{Policy Controller}
The policy controller component acts as an open loop controller that takes as input a series of desired goal positions. The main control occurs while iterating over the given desired goal positions. Iterating through desired goals, first, the environment is initialized with a reset function that returns the first state. The reset function is used as an input to the system identifier and desired goal. Next, in the main for loop of the controller, the policy is given $20$ timesteps to achieve the current desired goal with actions from the policy function. If the agent has achieved the current desired goal within the $1$ mm tolerance before $20$ timesteps, then a break is initiated and the next desired goal is set. This is outlined in Algorithm \ref{algo:policy_controller}.

\subsection{Single System Validation}
To validate our constraint-free, egocentric decay curriculum DeepRL method, we train different CTR systems and present error metrics for inverse kinematics and path following for each system. This was done to verify that the same method can be applied to various systems, resulting in an accurate learned policy. Additionally, we add state noise in the form of encoder and end-effector position noise. This is to demonstrate the learned policy is somewhat resilient to noise in the state. For simplicity, a $1^{\circ}$ standard deviation was selected. To determine the extension joint noise, a gear ratio of $0.001$ was used. For achieved goal tracking noise, a standard deviation $0.8$ mm based on an EM tracker (Aurora, NDI Inc., CA) precision data found in the documentation.

Secondly, to provide a pathway to hardware translation, we include initial results for domain randomization. In domain randomization, to transfer the policy from the source domain (simulation) to the target domain (hardware), a series of environmental parameters in the source domain is sampled from a randomized space. During training, the episodes are collected with the source domain with randomization sampling applied. This allows for the policy to be exposed to a variety of environmental parameters to generalize. In this way, the policy is trained to maximize the expected reward over a distribution of CTR configurations around the desired CTR configuration. Specifically, uniform domain randomization \cite{Tobin2017}, is implemented for the tube parameters specified in the simulation. These tube parameters include length, curved length, inner diameter, outer diameter, stiffness, torsional stiffness, and pre-curvature. In uniform domain randomization, an interval range from which the tube parameters are uniformly sampled should be defined. For example, for a randomization value of 5 \%, the inner diameter sampling of system 0, the lower range would be $0.7 - 0.05 \times 0.7 = 0.665$ and the upper range would be $0.7 + 0.05 \times 0.7 = 0.735$. The chosen CTR configuration for the domain randomization experiment was that of system $2$ with a domain randomization value of $\pm 5\%$ of each parameter.

\begin{table*}[tbph]
\centering
\caption{Four CTR systems with varied physical parameters. This table includes the lengths and curvatures.}
\label{tab:ctr_systems_lengths}
\begin{tabular}{@{}cccccccccc@{}}
\toprule
\multicolumn{1}{c}{\textbf{System}} &
  \multicolumn{1}{c}{\textbf{Tube}} &
  \textbf{\begin{tabular}[c]{@{}c@{}}Length\\ (mm)\end{tabular}} &
  \textbf{\begin{tabular}[c]{@{}c@{}}Curved\\ Length\\ (mm)\end{tabular}} &
  \textbf{\begin{tabular}[c]{@{}c@{}}Inner\\ Diameter\\ (mm)\end{tabular}} &
  \textbf{\begin{tabular}[c]{@{}c@{}}Outer\\ Diameter\\ (mm)\end{tabular}} &
  \textbf{\begin{tabular}[c]{@{}c@{}}Youngs\\Modulus\\ (GPa)\end{tabular}} &
  \textbf{\begin{tabular}[c]{@{}c@{}}Shear\\Modulus\\ (GPa)\end{tabular}} &
  \textbf{\begin{tabular}[c]{@{}c@{}}Precurvature\\ ($m^{-1}$)\end{tabular}} \\ \midrule
\multirow{3}{*}{System 0 \cite{khadem2020}}   & Inner & 431 & 103  & 0.7  & 1.1  & 102.5 & 187.9 & 21.3  \\
                                              & Middle & 332 & 113  & 1.4  & 1.8  & 685   & 115.3 & 13.1\\
                                              & Outer  & 174 & 134  & 2.0  & 2.4 & 169.6 & 142.5 & 3.5 \\ \midrule
\multirow{3}{*}{System 1 \cite{grassmann2018}}  & Inner & 370 & 45  & 0.3  & 0.4 &  500   & 230   & 15.8  \\
                                              & Middle  & 305 & 100  & 0.7  & 0.9 &  500   & 230   & 9.27 \\
                                              & Outer   & 170 & 100  & 1.2  & 1.5 & 500   & 230   & 4.37 \\ \midrule
\multicolumn{1}{l}{\multirow{3}{*}{System 2}} & Inner   & 309 & 145  & 0.7  & 1.1 & 75    & 25    & 1.68  \\
\multicolumn{1}{l}{}                          & Middle  & 275 & 114  & 1.4  & 1.8 & 75    & 25    & 11.6 \\
\multicolumn{1}{l}{}                          & Outer   & 173 & 173  & 1.83 & 2.39 & 75    & 25    & 10.8 \\ \midrule
\multirow{3}{*}{System 3 \cite{iyengar2020}}  & Inner   & 150 & 100  & 1.0  & 2.4 & 50    & 23    & 15.82 \\
                                              & Middle  & 100 & 21.6 & 3.0  & 3.8 & 50    & 23    & 11.8  \\
                                              & \multicolumn{1}{c}{Outer} & 70  & 8.8  & 4.4  & 5.4 & 50    & 23    & 20.04  \\ \bottomrule
\end{tabular}
\end{table*}

To evaluate this translated domain policy, we performed path-following tasks and inverse kinematics on system $2$ tube parameters. Although there exist more sophisticated methods of domain translation, as an initial work, we believe this demonstrates the feasibility to translate the policies trained to hardware. In $1000$ evaluation episodes, a mean error of $0.86$ mm was found with a standard deviation of $0.64$ mm. Using a straight-line path for testing for path following, the mean tracking error was $1.10$ mm with a standard deviation of $0.15$ mm. We compare these results to state-of-the-art in the next section. Without domain randomization, training results for inverse kinematics were summarized in Table \ref{tab:rotation_exps} under constraint-free experiments. For the path following, in a noise-free simulation, system $0$ had error metrics of $0.66 \pm 0.28$ mm for a helix path and $1.74 \pm 0.72$ mm in a noise-induced simulation. System $0$ is the longest system, with the highest errors, and was chosen for results, evaluation, and comparisons. 
\renewcommand{\algorithmicrequire}{\textbf{Input:}}
\renewcommand{\algorithmicensure}{\textbf{Output:}}
\begin{algorithm}[tb]
\caption{Policy Controller}
\begin{algorithmic}[1]
\Require{$G_{d}^0 \dots G_{d}^N$}
\Ensure{$A[\;]$ (Array of actions to achieve path)}
\Statex
\Function{Controller}{$G_{d} [\;]$}
    \For{$i \gets 1$ to $N$}
    \State $G_{current} \gets G_{d}[i]$
    \State $s_{current} \gets Reset(G_{d})$
        \For{$t \gets 0$ to $20$}
        \State $a_{current} \gets Policy(s_{current})$
        \State $s_{next}, r_{next}, terminal \gets Step(a_{current})$
        \State $A \gets A.append(a_{current})$
            \If{$terminal = True$}
            $break$
            \EndIf
        \EndFor
    \EndFor
    \State \Return {$A[\;]$}
\EndFunction
\end{algorithmic}
\label{algo:policy_controller}
\end{algorithm}%
\subsection{Generic Policy Validation}
To validate our generic policy method, we trained a generic policy for different combinations of two, three, and four CTR systems. For example, to generalize over two CTR systems, because we have four different CTR systems available to train on, we trained on all combinations of two systems resulting in a total of six experiments. Similarly, generalizing over three systems results in four experiments and a single experiment to generalize over all four CTR systems. Performing $1000$ evaluation episodes and summarizing the error metrics, the proposed method is able to generalize over multiple systems. The full error metrics are shown in Table \ref{tab:generic_system_exps}. Looking at the error metrics with respect to systems, there is a correlation between the length of the system and higher errors similar to the previous constrained and constraint-free experiments. System $0$ is the longest length, and consistently has the largest error metrics. We believe this is because of the workspace size, as overall length increases, the agent will require more training steps and experiences. Another factor is that in general, longer CTR systems have larger errors, and thus comparisons are done with the percentage of robot length. To mitigate this effect, a sampling strategy is used where the sampling of the system used in the environment is based on the lengths of the systems. In this length-based sampling strategy, the categorical distribution is proportionate to the length of each system. Each system probability is the system length divided by the sum of the system lengths being generalized. In this manner, systems that are longer and that have larger workspaces are sampled more during training and have more experiences for the policy to train. Evaluating this sampling strategy for generalizing over four CTR systems, the error metrics were improved from the previous uniform sampling. To validate the generalization, we performed a helix path following task with system $0$ as seen in Fig. \ref{fig:sample_gen_helix}. The error metrics were a mean tracking error of $1.01$ mm and a standard deviation of $0.41$ mm with $50$ desired goal points in the path. Performing a noise-induced experiment the error metrics were a mean tracking error of $1.86$ mm with a standard deviation of $0.8$ mm. When the number of points was increased to $100$, error metrics were a mean tracking error of $0.91$ mm with a standard deviation of $0.41$ mm. In a noise-induced simulation, the mean tracking error was $1.91$ mm with a standard deviation of $0.78$ mm. 

To compare our generic egocentric constraint-free method, we also train using proprioceptive representation for a four-system generic policy with constrained rotation and one with constraint-free rotation to compare to our egocentric constraint-free policy. We summarize the results in Table \ref{tab:generic_comp}. To note is the importance of removing rotation constraints as seen in the mean and standard deviation of errors from constrained proprioceptive to constraint-free proprioceptive. Error metrics are greatly reduced with this improvement. Finally, using an egocentric representation does improve metrics but has less of a significant impact as compared to rotational constraints.
\begin{table}[tb]
\centering
\caption{Experimental comparisons of rotational and joint representation per system.}
\label{tab:generic_comp}
\begin{tabular}{@{}ccccc@{}}
\toprule
\textbf{Experiment}                                                                   & \textbf{\begin{tabular}[c]{@{}c@{}}First\\ System\end{tabular}} & \textbf{\begin{tabular}[c]{@{}c@{}}Second\\ System\end{tabular}} & \textbf{\begin{tabular}[c]{@{}c@{}}Third\\ System\end{tabular}} & \textbf{\begin{tabular}[c]{@{}c@{}}Fourth\\ System\end{tabular}} \\ \midrule
\textbf{\begin{tabular}[c]{@{}c@{}}Constrained\\ Proprio-\\ ceptive\end{tabular}}     & \begin{tabular}[c]{@{}c@{}}$4.68 \pm$\\  $15.96$\end{tabular}   & \begin{tabular}[c]{@{}c@{}}$1.98 \pm$\\ $8.45$\end{tabular}      & \begin{tabular}[c]{@{}c@{}}$4.91 \pm$\\ $24.05$\end{tabular}    & \begin{tabular}[c]{@{}c@{}}$0.87 \pm$\\ $1.56$\end{tabular}      \\ \midrule
\textbf{\begin{tabular}[c]{@{}c@{}}Constraint-free\\ Proprio-\\ ceptive\end{tabular}} & \begin{tabular}[c]{@{}c@{}}$0.77 \pm$\\ $0.37$\end{tabular}     & \begin{tabular}[c]{@{}c@{}}$0.80 \pm$\\ $2.63$\end{tabular}      & \begin{tabular}[c]{@{}c@{}}$0.78 \pm$\\ $0.34$\end{tabular}     & \begin{tabular}[c]{@{}c@{}}$0.73 \pm$\\ $0.31$\end{tabular}      \\ \midrule
\textbf{\begin{tabular}[c]{@{}c@{}}Constraint-free\\ Egocentric\end{tabular}}         & \begin{tabular}[c]{@{}c@{}}$0.75 \pm$\\ $0.29$\end{tabular}     & \begin{tabular}[c]{@{}c@{}}$0.72 \pm$\\ $0.26$\end{tabular}      & \begin{tabular}[c]{@{}c@{}}$0.73 \pm$\\ $0.26$\end{tabular}     & \begin{tabular}[c]{@{}c@{}}$0.63 \pm$\\ $0.23$\end{tabular}      \\ \bottomrule
\end{tabular}
\end{table}
\subsection{Comparisons to the State of the Art}
\begin{figure}[tb]
    \centering
    \subfloat[\label{fig:jac_circle}]{\includegraphics[width=0.5\linewidth]{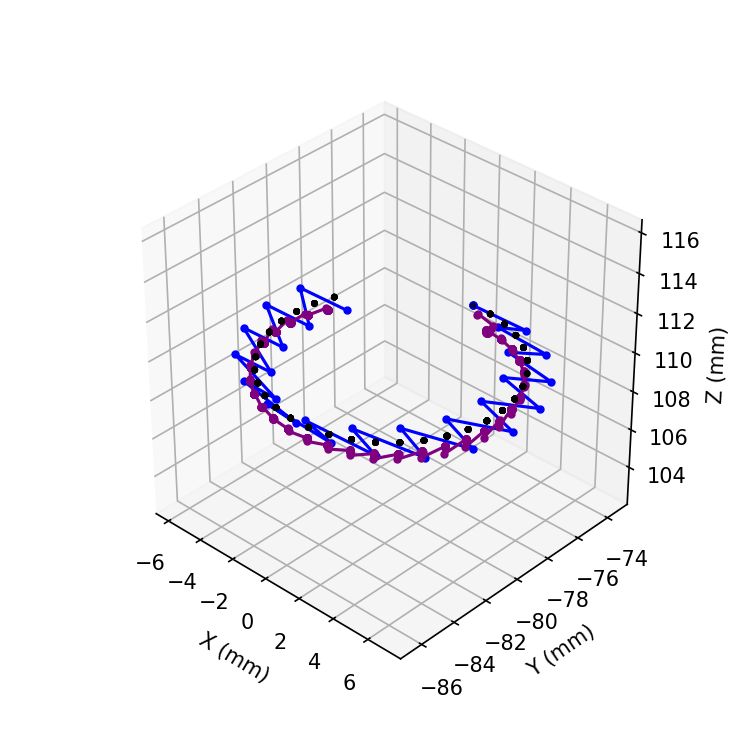}}%
    \subfloat[\label{fig:jac_line}]{\includegraphics[width=0.5\linewidth]{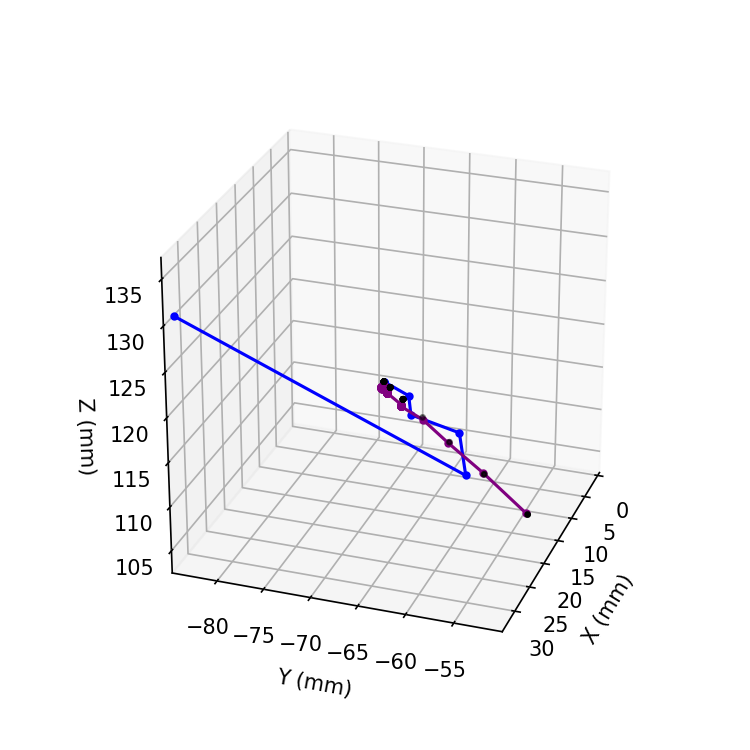}}%
    \caption{System $0$ following a circular path \protect\subref{fig:jac_circle} and linear path \protect\subref{fig:jac_line} comparing Jacobian (blue) methods to our Deep RL method (purple) for path following. \label{fig:jac_comp}}
    \label{fig:jac_rl_comp}
\end{figure}

To compare our inverse kinematics and path following results to a previous state-of-the-art classical methods and deep learning methods, we convert errors to a percentage of robot length. First, we present inverse kinematics results for our constraint-free egocentric decay for each system when trained separately ie. not the generic policy. Mean and standard deviation as percentage for each system was $0.16\% \pm 0.065\%$, $0.17\% \pm 0.18\%$, $0.20\% \pm 0.08\%$ and $0.3\% \pm 0.13\%$.  When taken as a percentage of the robot length, the similarity of the error metrics is noteworthy. In our generalization method, for the four system generalization inverse kinematics for each system is $0.18\% \pm 0.02\%$, $0.19\% \pm 0.02\%$, $0.24\% \pm 0.02\%$ and $0.42\% \pm 0.04\%$. For tip tracking with the more complex helix path errors were $0.23\% \pm 0.095\%$ for system 0 with $50$ goal points. Increasing the number of desired goal points to $100$, the following mean tracking errors of $0.20\% \pm 0.08\%$  as seen in Fig. \ref{fig:sample_gen_helix}.
As reported in \S \ref{sec:related_work}, Jacobian-based methods can achieve errors of $0.5\%$ to $0.9\%$. As shown in Fig. \ref{fig:jac_comp}, our DeepRL method is able to avoid joint limits, especially in extension, whereas the Jacobian approach does not include joint limits in the linearization. This comparison was done on system 0 with $K_p = 2I$ for a linear and circular path. The error metrics found were  $1.15$ mm $\pm$ $0.32$ for the Jacobian method and $0.62$ mm $\pm$ $0.07$ for our DeepRL method in Fig \ref{fig:jac_circle}. More importantly, the Jacobian-based method was unable to follow some circular, linear, and helical trajectories that our DeepRL method successfully completed due to the Jacobian not including joint limits in the formulation, even if the damped-least squares method was used with $\Lambda = 0.45$. as seen in Fig. \ref{fig:jac_line}. The advanced MPC method can achieve tip errors of $0.3\%$ to $0.5\%$, however, we were unable to perform comparisons in simulation as the open-source simulation code was for a two-tube system. Our method does perform comparably to the reported state-of-the-art in simulation, however, this work is only in simulation and does not include constraints for snapping conditions in the model used. The approach will need to be validated in hardware. One possible way to consider snapping and singularity is to design a dense reward function that includes the minimization of elastic energy, hence avoiding snapping conditions. For our domain randomization results, as a percentage of robot length, the inverse kinematics errors were $0.86\% \pm 0.21\%$. Following a straight-line path, the errors were $0.36\% \pm 0.05\%$. The domain randomization metrics are higher, however, the aim is to demonstrate the feasibility of the transfer method. To validate, we will need to compare transfer to hardware with and without domain randomization or other transfer methods.

\section{Conclusion}
\label{sec:conclusion}
In this work, we investigate DeepRL, an end-to-end method for kinematic control of CTRs. Specifically, we aim to explore constraints on tube rotation and the impact on error metrics. Furthermore, the first proof-of-concept in system generalization is developed, not yet done for any deep learning method for CTRs. Finally, to provide initial work for hardware, we provided an initial pathway for domain transfer or policy transfer from simulation to hardware using Sim2Real domain randomization. Our method does demonstrate error metrics that perform well, however, validation in hardware is needed. Moreover, other domain transfer methods should be explored. Unaddressed in this work is the issue of snapping, whereby torsion in the tubes causes a rapid transition to a lower energy state resulting in a snapping motion. By creating simulation frameworks that utilize Dynamics models \cite{sadati2022}, snapping configurations could be avoided through rewards. We believe, with this work, we have demonstrated that deepRL methods may be able to outperform model-based methods for inverse kinematics and control for CTRs, similar to deep learning methods for forward kinematics and shape estimation.





\ifCLASSOPTIONcaptionsoff
  \newpage
\fi



%
\bibliographystyle{IEEEtran}
\bibliography{refs}



%
\begin{IEEEbiographynophoto}{Keshav Iyengar}
Keshav Iyengar (Student Member, IEEE) is currently a Ph.D. student at University College London in Computer Science under the Welcome / EPSRC Centre for Interventional and Surgical Sciences (WEISS). He received his BASc in Mechanical Engineering at the University of Waterloo, Canada in 2012 and his MSc in Robotics and Computation at University College London in 2018. His research focus is on model-free control of continuum robots for minimally invasive surgery.
\end{IEEEbiographynophoto}
\begin{IEEEbiographynophoto}{Sarah Spurgeon}
Sarah Spurgeon (F’20) received her B.Sc. and D.Phil. degrees from the University of York, UK in 1985 and 1988, respectively, and is now Professor of Control Engineering at UCL, UK. Sarah’s research interests are in the area of systems modeling and analysis, robust control, and estimation in which areas she has published over 300 research papers. She was awarded the Honeywell International Medal for ‘distinguished contribution as a control and measurement technologist to developing the theory of control’ in 2010 and an IEEE Millennium Medal in 2000. She is currently Editor in Chief of IEEE Press and Vice President of Publications of IFAC.
\end{IEEEbiographynophoto}
\begin{IEEEbiographynophoto}{Danail Stoyanov}
Dan Stoyanov is a Professor at UCL Computer Science holding a Royal Academy of Engineering Chair in Emerging Technologies and serving as Director of the Wellcome/EPSRC Centre for Interventional and Surgical Sciences (WEISS). He graduated from King’s College London before completing his PhD at Imperial College London. His research interests are focused on surgical robotics, surgical data science, and the development of surgical AI systems for clinical use. 
\end{IEEEbiographynophoto}





\end{document}